\pdfoutput=1

\documentclass[11pt]{article}

\usepackage{CJKutf8}
\usepackage{ACL2023}

\usepackage{times}
\usepackage{latexsym}
\usepackage{soul}
\usepackage[T1]{fontenc}

\usepackage[utf8]{inputenc}

\usepackage{microtype}

\usepackage{inconsolata}

\usepackage{xspace}
\usepackage{graphicx}
\usepackage{amssymb}
\usepackage{multirow}
\usepackage{multicol}
\usepackage{array}
\usepackage{booktabs}
\usepackage{algorithm}
\usepackage{algorithmic}
\usepackage{amsmath}
\usepackage{makecell}
\usepackage{amsfonts}
\usepackage{bbold}

\title{A New Dialogue Response Generation Agent for Large Language Models\\ by Asking Questions to Detect User’s Intentions }

\author{Siwei Wu, Xiangqing Shen, and Rui Xia \\
        School of Computer Science and Engineering, \\ Nanjing University of Science and Technology, China \\
        \texttt{\{wusiwei, xiangqing.shen, rxia\}@njust.edu.cn}}

\begin{document}
\maketitle
\begin{abstract}

Large Language Models (LLMs), such as ChatGPT, have recently been applied to various NLP tasks due to its open-domain generation capabilities. 
However, there are two issues with applying LLMs to dialogue tasks. 1. During the dialogue process, users may have implicit intentions that might be overlooked by LLMs. Consequently, generated responses couldn't align with the user's intentions.
2. It is unlikely for LLMs to encompass all fields comprehensively. In certain specific domains, their knowledge may be incomplete, and LLMs cannot update the latest knowledge in real-time.
To tackle these issues, we propose a framework~\emph{using LLM to \textbf{E}nhance dialogue response generation by asking questions to \textbf{D}etect user's \textbf{I}mplicit in\textbf{T}entions} (\textbf{EDIT}). 
Firstly, EDIT generates open questions related to the dialogue context as the potential user's intention; Then, EDIT answers those questions by interacting with LLMs and searching in domain-specific knowledge bases respectively, and use LLMs to choose the proper answers to questions as extra knowledge; Finally, EDIT enhances response generation by explicitly integrating those extra knowledge. 
Besides, previous question generation works only focus on asking questions with answers in context. In order to ask open questions, we construct a Context-Open-Question (COQ) dataset.
On two task-oriented dialogue tasks (Wizard of Wikipedia and Holl-E), EDIT outperformed other LLMs.

\end{abstract}

\begin{figure*}[!htp]
    \centering
    \includegraphics[width=0.99\textwidth]{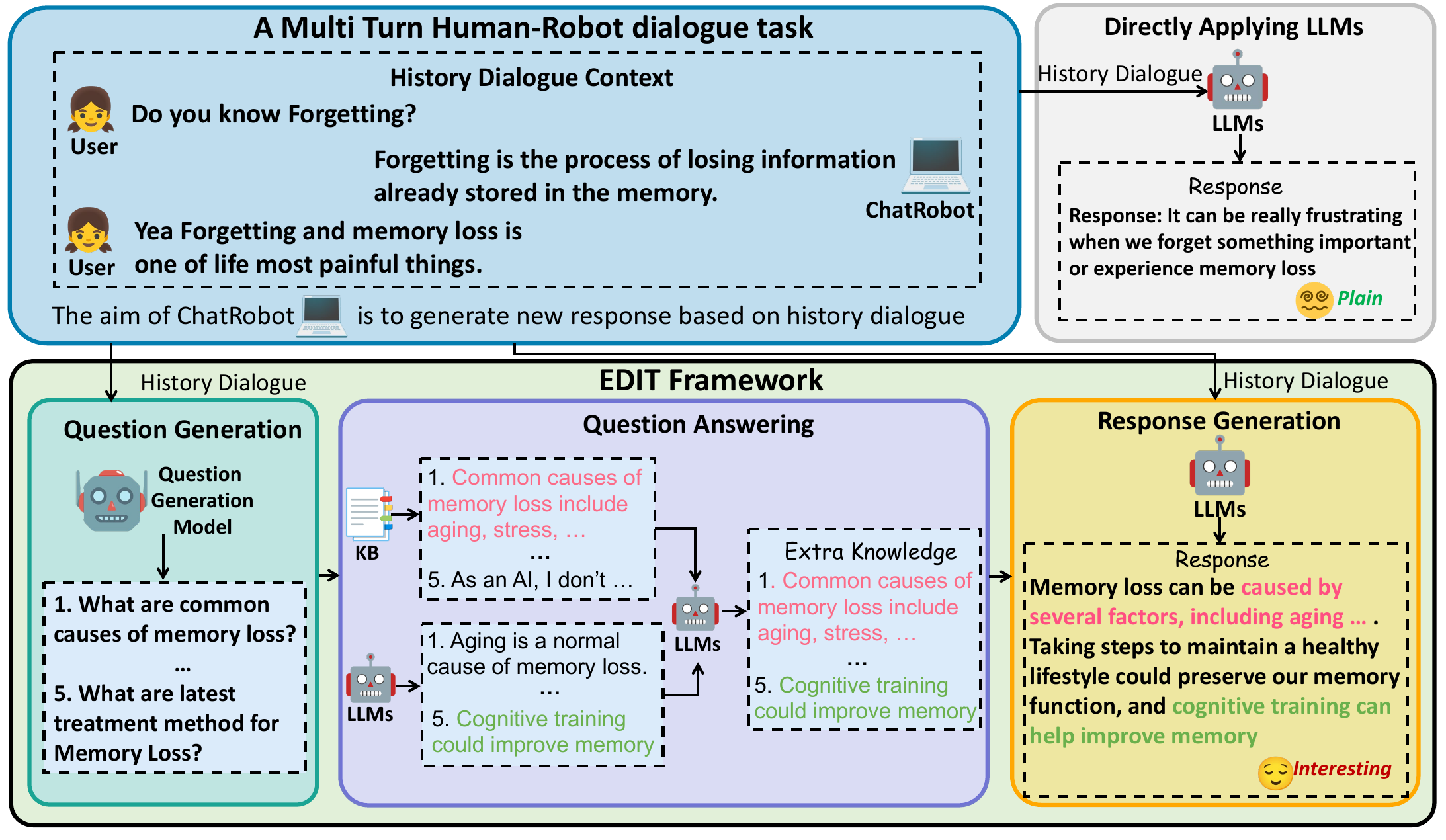}
    \caption{The framework of EDIT. The upper left part is an example of multi turn human-robot dialogue task, and , in the next turn, the ChatRobot needs to generate a response based on the historical dialogue context. The upper right part represents the method directly applying LLMs to generate response. And the lower part is the illusion of our EDIT method. It consists of three modules: Question Generation, Question Answering and Response Generation.
    }
    \label{fig:AQRA}
\end{figure*}

\section{Introduction}

Large language models (LLMs) like ChatGPT\footnote{\url{https://chat.openai.com}} and LLaMA~\cite{DBLP:journals/corr/abs-2302-13971} have recently demonstrated remarkable performance on various Natural Language Processing (NLP) tasks~\cite{DBLP:journals/corr/abs-2302-04023}, such as commonsense reasoning~\cite{DBLP:journals/corr/abs-2303-16421}, sentiment analysis~\cite{DBLP:journals/corr/abs-2304-04339}, and recommendation system~\cite{DBLP:journals/corr/abs-2305-13112}, which makes applying LLMs to domain-specific generative tasks attracting extensive attention~\cite{DBLP:journals/corr/abs-2303-17580}. 

In dialogue tasks, LLMs generally excel at understanding the semantics of the context and providing relevant responses. However, there are a few instances where LLMs generate plain responses that lack sufficient information and may not fully meet the user's needs.

As depicted in Figure~\ref{fig:AQRA}, when presented with historical dialogues context "LLM: Forgetting is the process of losing information already stored in the memory. User: Yea Forgetting and memory loss is one of life most painful things.", the LLM only acknowledges the context by expressing that "Memory Loss is frustrating". However, it fails to provide additional details that users might find interesting, such as the causes and possible treatments for "Memory Loss."

There are two main reasons for this:

1. During conversations, LLMs may overlook implicit intentions of users. LLMs are trained on a vast amount of data and implicitly contain a considerable amount of commonsense knowledge in a parameterized method. Additionally, instruction-tuning methods enables LLMs to follow instructions and utilize their own knowledge to solve problems. However, these instructions typically use the explicit intent shown in the natural text.
As the dialogue progresses, users will have many implicit intentions hidden in the conversation. These implicit intentions sometimes provide valuable insights for generating responses. While, due to the lack of understanding of these intentions, LLMs tend to prioritize dialogue coherence rather than utilizing relevant common sense knowledge related to users' implicit intentions.
As shown in Figure~\ref{fig:AQRA}, when user says "Yes, forgetting and memory loss is one of life's most frustrating things.", her implicit intention behind this statement might be the hope that LLMs can provide ways to prevent forgetting. As for the question "What are the common causes of memory loss?", LLMs can provide relevant basic knowledge to prevent forgetting since they possess this information. However, without understanding the user's intention, LLMs fail to employ this knowledge in generating responses.

2. Although LLMs possess a vast amount of knowledge, it is not all-encompassing and may not cover every field comprehensively. In a few specific domains, its knowledge may also be incomplete, and LLMs cannot update the latest knowledge in real-time. These limitations may hinder the ability of LLMs to generate responses that align with the user's implicit intentions, as demonstrated in Figure~\ref{fig:AQRA}. For instance, when the LLM is asked the question "What are latest treatment method for Memory Loss?", it may not possess the necessary up-to-date knowledge in this particular field, thereby rendering it incapable of providing an answer to this question.

In order to solve the problem of generating plain responses by LLMs and increase the informational content, we propose a framework using LLM to enhance dialogue response generation by asking questions to detect user's implicit intentions (\textbf{EDIT}). This framework consists of three main modules: Question Generation, Question Answering, and Response Generation.
Firstly, the Question Generation Module trains a Question Generation Model to generate a series of questions based on the input historical dialogue context, which serves as potential user's implicit intentions. This aims to enable the LLMs to better understand users' implicit intentions.
Subsequently, on the one hand, the Question Answering Module allows LLMs to answer these generated questions, thereby exploring relevant knowledge within the LLMs; on the other hand, we have constructed a domain-specific knowledge base for different downstream dialogue tasks, and Question Answering Module calculates the semantic similarity between questions and knowledge in KB, searching for sentence-level and question-related information to provide as answers.
In addition, we have sent two types of answers to questions into LLMs to determine which answer is more appropriate and regard the final answer as extra knowledge.
Finally, the Question Answering Module inputs dialogue context and extra knowledge related to LLMs, utilizing this knowledge to generate new responses.

In the past, question generation tasks often ignore addressing the implicit intention detection needs of users. Question generation task is primarily used as a supplementary task to answer questions about the explicit text in the given context.
The implicit intention of the user may not be explicitly stated in the context and it may be closely related to some extra knowledge. To tackle this challenge, EDIT framework trains a Question Generation Model to generate context-related questions and regards those questions as potential user's intentions.
To enable the EDIT framework to generate such questions, we create a Context-Open-Question dataset. We further evaluate the Question Generation Model trained on this dataset through both automatic and manual assessments.
Compared to other LLMs, our Question Generation Model demonstrates superior ability in generating questions that are relevant to knowledge beyond the given context.

We conduct a performance comparison between EDIT and other LLMs in terms of generating responses. Additionally, we perform human evaluations and utilize LLM (GPT4) to evaluate the results.
In comparison to other LLMs, EDIT has achieved SOTA performance in downstream tasks. This serves as evidence of the effectiveness of EDIT framework.
Furthermore, we carry out ablation experiments to verify that integrating LLM to answer questions and utilizing a knowledge base for retrieving answer questions can indeed enhance overall performance.

\section{Related Works}

\paragraph{LLMs Application.}
As LLMs, like ChatGPT, have demonstrated exceptional performance in various natural language processing (NLP) tasks, there has been a growing interest in exploring their applications. These models have found direct applications in several domains, including Healthcare\cite{DBLP:journals/corr/abs-2303-04360,DBLP:journals/corr/abs-2301-10035,DBLP:journals/corr/abs-2304-03347}, Education\cite{DBLP:conf/iticse/MalinkaPFHJ23,DBLP:journals/corr/abs-2305-03433,DBLP:journals/corr/abs-2305-18303}, Finance\cite{DBLP:journals/corr/abs-2303-17564} and Scientific research\cite{DBLP:journals/corr/abs-2303-18223}.
Besides, there are also some works to realize the application of LLM on specific tasks through Agent.
The study by \cite{DBLP:journals/corr/abs-2308-03427} implements the assessment of language models' ability to perform reasoning processes through an Agent. Meanwhile, \cite{DBLP:journals/corr/abs-2308-02151} focuses on optimizing the reasoning and planning capabilities of surrogate models using reinforcement learning techniques. Additionally, \cite{DBLP:journals/corr/abs-2308-01734} harnesses the story generation capabilities of large language models and applies them to the realm of imaginative play.
Furthermore, there are various practical applications that leverage Large Language Models (LLMs) in real-life scenarios, including a series of plug-ins \cite{DBLP:journals/corr/abs-2303-18223} designed for Copilot and OpenAI. While these applications can successfully address specific tasks through agents, they still grapple with the challenge that LLMs may ignore users' implicit intentions and are unlikely to cover all fields.
In light of these considerations, our work introduces the EDIT framework, aimed at asking questions to detect users' potential intentions and use LLMs and related knowledge to generate responses with more information.

\paragraph{Question Generation.}

Before the advent of LLMs, contextual information in different tasks was primarily used as a supplementary source of information for task-related questions. In many cases, the answers to most task questions could be obtained either by leveraging context assistance or by finding the relevant information within the given context.
Question generation usually involves generating questions based on a given context. Depending on the type of context, there are different levels of text input generation problems. For example, document-level text input generation problems (Pan et al., 2020), paragraph-level text input generation problems (Zhao et al., 2018), and sentence-level text input generation problems (Gao et al., 2019).
Answering questions can be helpful in the question generation process. In the answer span type question generation task, the answer to the question is a fragment of the contextual text (Rajpurkar et al., 2016). On the other hand, abstractive answers require summarizing the provided question-answer pairs within the context (Nguyen et al., 2016). Additionally, answer-agnostic QG tasks only provide the context without specific answer information (Chen et al., 2018).
Most question generation (QG) tasks are used as a supplement to the question answering (QA) task. The generated questions usually rely on the information present in the given context, and they do not involve mining information from external sources. 
However, users' implicit intentions may some question beyond context. Ans it is important to consider mining information outside the given context. As part of our work, we have developed a Context-Open-Question dataset that focuses on generating questions about out-of-context information. This dataset is used to train the Question Generation Model for generating questions.

\section{Approach}

As illustrated in Figre~\ref{fig:AQRA}, our EDIT framework consists of three modules: Question Generation Module, Question Answering Module, and Response Generation Module. Firstly, the Question Generation Module generates a series of related open questions according to the dialogue context and regards these questions as the user's potential implicit intentions. The Question Answering Module obtains the answers to the questions by interacting with LLM and retrieving domain-specific knowledge bases, and it also lets LLM choose proper answer between these answers as extra knowledge.
Finally, the Response Generation Module inputs the extra knowledge and historical dialogue context to LLMs to generate a new response.

\subsection{LLM for Dialogue Task} 
\begin{CJK}{UTF8}{gbsn}
\end{CJK}
We denote the dialogue history of a conversation $C=\{u_1,r_1,...,u_t\}$ at time step $t$, where $u$ and $r$ represent the utterances from the user and the chatbot respectively. The $ExtraKnow$ denotes extra context-related knowledge. The purpose of our task is to predict the chatbot's responses $r_{t+1}$ by using extra knowledge $ExtraKnow$ and dialogue history context $C$.

\subsection{Question Generation}
To determine what extra knowledge can be used to assist in generating responses, we train a Question Generation Model that is capable of generating context-related questions and we regard these questions as user's implicit intentions. To accomplish it, we create a Context-Open-Question (COQ) dataset.
\subsubsection{Contex-Open-Question Dataset Creation}
\label{dataset}
\begin{figure}[!htp]
\centering
\includegraphics[width=0.99\columnwidth]{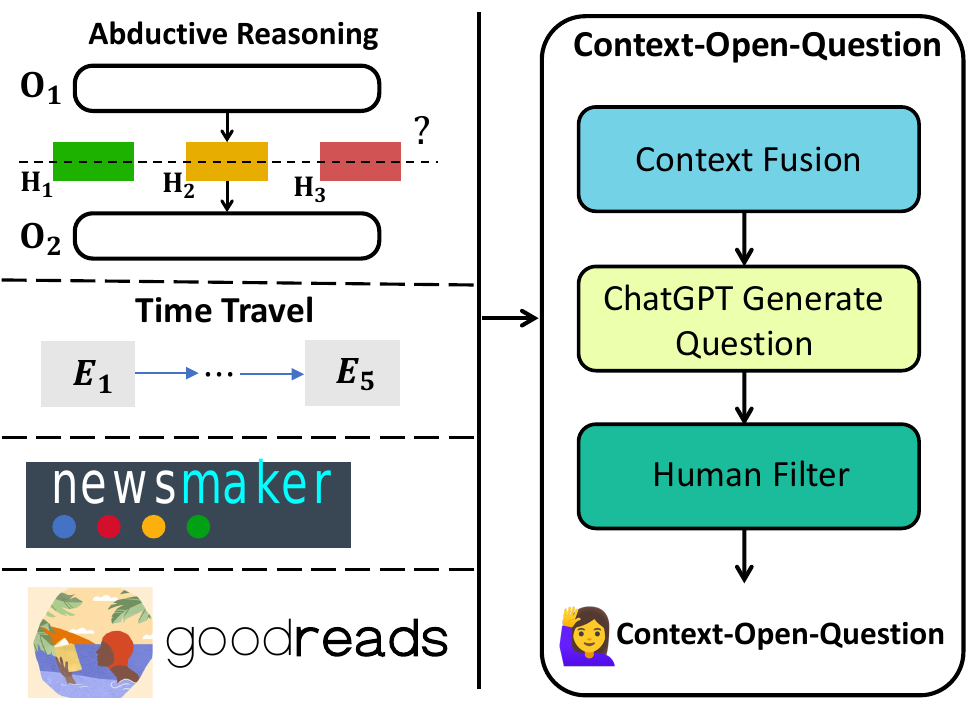}
\caption{The process of COQ dataset curated.}
\label{figure: CQ Curated}
\end{figure}

As shown in Figure~\ref{figure: CQ Curated}, to enable the Question Generation Model of EDIT framework to ask open questions for dialogue context in various real scenarios, we developed the Contex-Open-Question dataset. We train a Question Generation Model using the COQ datasets.
In creating the dataset, we employ a method combining ChatGPT and manual annotation. This approach allows us to create high-quality data that meets the requirements of our EDIT framework.

\paragraph{Context Datasets Selection.}
\label{Context Datasets Choosing}

We carefully selected four datasets that represent different daily life scenarios. These datasets were merged to create contextual datasets.

Goodreads Reviews\footnote{\url{https://mengtingwan.github.io/data/goodreads.html}} has constructed a comprehensive book review dataset to address real-life large-scale application scenarios. This dataset consists of 1,378,033 English book reviews collected from 25,475 books and encompasses the review data provided by 18,892 users.
Review data, being the most prevalent form of textual information in our daily lives, holds immense significance. Book reviews, in particular, offer a wealth of storytelling content that can be associated with various information beyond the review context text.

News Category~\cite{misra2022news} extracted about 210,000 news headlines from HuffPost in 2012-2022 as high-quality real news sources to address the issue of the overwhelming amount of fake news in today's world. Compared with the more colloquial text forms on various social media, news texts are very formal. It can provide a different environment to question.
Abductive Commonsense Reasoning~\cite{DBLP:conf/iclr/BhagavatulaBMSH20} is inference to the most plausible explanation. It involves inferring intermediate process events based on two causally related events. This type of discontinuous causal event context is frequently encountered in our daily lives and requires the utilization of a substantial amount of external commonsense knowledge for effective reasoning.

Time Travel~\cite{DBLP:conf/emnlp/QinBHBCC19} offers a narrative experience encompassed within five sentences, seamlessly unfolding in continuous time. In our daily lives, most events we encounter are comprised of multiple consecutive moments, and this dataset effectively simulates various real-life situations. It captures the essence of everyday life, presenting a rich tapestry of interconnected events.

We combined the context from those four datasets to create the Context of the Context-Open-Question dataset.

\paragraph{Open Question Annotation}

To enable the EDIT model to generate context-related questions about diverse extra knowledge, we initially allow ChatGPT to generate questions based on context through the Prompt "Give you a context: \{Context\}. Help me ask questions, which is unrelated to the person in the context ... ".     
Afterward, we manually filter the generated Context-Open-Question pairs to ensure that the questions are relevant to the context. Additionally, we make sure that the questions can be handled by LLMs.

\begin{CJK}{UTF8}{gbsn}
\begin{table}[!htp]
\centering
\scriptsize
\small
\resizebox{0.8\columnwidth}{!}{
    \begin{tabular}{rcccc}
    \hline
    \multicolumn{1}{l}{}                    &                     & Train                 & Test                  & Valid                 \\ \hline
    \multirow{4}{*}{COQ}                     & ACR & 1564                  & 108                   & 115                   \\
                                            & TT         & 1334                  & 78                    & 83                    \\
                                            & NC       & 2291                  & 197                   & 195                   \\
                                            & GR   & 689                   & 34                    & 42                    \\ \hline
    \end{tabular}
}
\caption{The COQ dataset statistics. The COQ stands for Context-Open-Question dataset. The ACR, TT, NC, GR, and WoW represent the five datasets of Abductive Commonsense Reasoning, Time Travel, News Category, Goodreads Reviews, and Wizard of Wikipedia, respectively.}

\label{tab: Staistics of CQCK}
\end{table}
\end{CJK}

After manual filtering, we constructed a total of 5878 Context-Open-Quesion (COQ) training data; most of the four data sets selected can still retain more than 1300 training data after manual filtering. The GR dataset contains numerous samples that allow for discussions on the subject matter and content of books. And ChatGPT makes it easier to generate subjective questions related to these book contents and these subjective questions will be filtered.
As a result, there are only 689 pieces of training data left in the GR dataset.

\subsubsection{Question Generation Model}

To enable EDIT framework to generate various context-related open questions, we train a Question Generation Model to achieve it based on the Context-Open-Question data.

The input context is denoted as $C$. For each context $C$, there are multiple corresponding questions ${q_{1},...,q_{n}}$.  We concatenate all the questions together to form a sequence denoted as $$Q=[q_{1},...,q_{n}].$$

To generate the questions, we employ a generative model that generates a sequence $Q=[y_1,...,y_T]$ of length $T$ given the input context $C$.
The generation probability of each token $y_t$ is conditioned on the previously generated tokens $y_{<t}$ and the input context $C$, represented as:
$$P(y_t|y_{<t},C)=\text{GeneraiveModel}(y_{<t},C),$$

We use the standard negative log-likelihood (NLL) loss on the target question sequence $Q$ to train the model. The loss is computed as:
$$Loss = -\sum_{t=1}^{T}logP(y_t|y_{<t},C),$$

\subsection{Question Answering}

As illustrated in the right part of Figure \ref{fig:AQRA}, the Question Answering part of EDIT consists of three components: Answer Generation by LLMs, Retrieving Answers in KB, and Answer Integration.
LLMs possess a significant amount of implicit knowledge. However, in a few instances, it would ignore the user's implicit intentions and may not incorporate the relevant knowledge to generate response. So the Answer Generation Module answers questions by interacting with LLMs to mine implicit knowledge in LLMs.
In addition, LLMs are unlikely to fully understand the knowledge in all fields and cannot update it in real-time. The Retrieving Answers in KB part makes up for this by searching the answers in the Domian-Specific Knowledge Base.
Finally, Answer Integration choose the final answer to question based on these two kinds of answers.

\subsubsection{Answer Generation by LLMs}

Although LLMs possess a vast amount of commonsense knowledge, it may ignore the users' implicit intention to use relevant knowledge. As a result, LLMs mainly focus on contextual semantic coherence. In our approach, we guided LLM to answer open questions generated by the Question Generation Module to mine implicit knowledge in it:
$$answer_{LLM}=LLM(q),$$
where $q$ is the context-related open question generated by Question Generation Module.

And designed Prompt is: "Give you a question: {question}, Please answer it as briefly as possible.".

\subsubsection{Retrieving Answer in Knowledge Base}

LLMs unlikey fully understand the knowledge in all fields and cannot update it in real-time. As a result, there are certain questions that they cannot be able to answer. To overcome this limitation, we retrieve answers from a Domain-Specific Knowledge Base to get the human-summarized supplementary knowledge.
Besides, for each downstream task, we have developed a dedicated Domain-Specific Knowledge Base that aligns with the required knowledge. And in different downstream tasks, we use corresponding Domain-Specific Knowledge Base to retrieve answers.

\paragraph{Domain-Specific Knowledge Base Creation.}
For each sample in our chosen downstream task-oriented tasks, Holl-E and Wizard of Wikipedia, it provides the associated knowledge document. 
We split all those knowledge documents into sentences and built a Domain Specific Knowledge Base for each downstream task.
\paragraph{Answering Question in Domain-Specific Knowledge Base.}
\begin{CJK}{UTF8}{gbsn}
\end{CJK}
SentenceBERT\cite{DBLP:conf/emnlp/ReimersG19} to calculate semantic similarity to search for answers to questions in Domain-Specific KB.

For each question $q$ generated by the Question Generation Module, we send it to SentenceBERT, and pass the final layer's output through Meanpooling to obtain the question's representation $e_{q}$.
We also use SentenceBERT to obtain the representation $e_k$ of each knowledge sentence $k$ in the KB, and regard representation matrix of all knowledge representations as $E_k$.

We calculate the semantic similarity (cosine similarity) between question $q$ and knowledge $k$ as score of regarding this knowledge as answer of question:
$$S=CosSim(e_q,E_k).$$
We select the knowledge ${k_1,k_2,...k_L}$ with the top $L$ score as the knowledge related to the problem.
In our experiments, we set $L=10$.

Besides, this knowledge is discrete, it cannot form a semantically continuous natural language text. Therby, we use LLM to organize this knowledge as the answer to the question:

$$answer_{KB}=LLM(k_1,k_2,...k_L).$$

The designed Prompt is: "Give you a question: \{prompt\}, Please answer it use those knowledge: ${k_1,k_2,...k_L}$.".

\subsubsection{Answer Integration}
LLMs would refuse to answer some extreme questions and knowledge they do not understand; The answers to questions obtained through retrieving in knowledge base are often one-sided and cannot answer the questions well.
We use LLMs to integrate those two kinds of answers: 
$$Answer = LLM(q,answer_{LLM},answer_{KB}).$$

And the designed prompt is "Give you a question: {q}, and two answers to it, AnswerA: {answerLLM}, AnswerB:{answerKB}, please tell me which is better?".

For contact answers to all questions related to a context as the extra knowledge:
$$ExtraKnow=[Answer_1,...,Answer_n],$$
where $Answer_i$ is the answer of question $q_i$, $n$ the the number of questions, and $[\cdot]$ represents text contact.

\subsection{Response Generation}

While LLMs excel at generating conversational responses, in some instances, they may ignore the users' implicit intentions and use related knowledge to generate responses. To address this limitation and enable LLMs to provide responses with more valuable information, we propose utilizing context-related knowledge generated by EDIT as supplementary input for LLMs. By incorporating historical dialogue context $C$ and extra context-related knowledge $ExtraKnown$, LLMs can better generate responses in dialogue-based tasks:
$$R=LLM(C,ExtraKnown).$$

Specifically, we use the prompt: "Give you a context: \{context\} and some knowledge \{knowledge\}. Please use those knowledge to just generate next response of \{next person\}."
The output of LLMs is the generated response $R$ of our EDIT framework.

\begin{table*}
\begin{center} 
\footnotesize
\resizebox{1.8\columnwidth}{!}{
    \begin{tabular}{l|cc|cc} 
    \hline
    \multicolumn{1}{c|}{~} & \multicolumn{2}{c|}{\textbf{Wizard of Wikipedia}} & \multicolumn{2}{c}{\textbf{Holl-E}} \\ 
    LLMs & {\sc Reasonable} & {\sc GPT4} & {\sc Reasonable} & {\sc GPT4} \\
    \hline
    text-davinci-001 & 33.00 & 72.00 &  40.00 & 67.00  \\
    text-davinci-002 & 43.00 & 58.50 & 53.00 & 56.50  \\
    text-davinci-003 & 45.00 & 61.00 & 56.19 & 55.50  \\
    \hline
    ChatGPT & 72.50 & 84.50 & 74.29 & 85.50  \\
    EDIT(ChatGPT) & \textbf{93.50} & \textbf{85.50} & \textbf{87.62} & \textbf{87.00} \\
    -w/o KB's Answer & 91.50 & - & 86.19 & - \\
    -w/o ChatGPT's Answer & 82.38 & - & 84.00 & - \\
    \hline
    \end{tabular}
}
\end{center}
\caption{The response results on downtown stream task. We report the Human Evaluation metric (REASONABLE) and the Automatical Model Evaluation metric (GPT4).
}
\label{tab:Main Result}
\end{table*}

\section{Experimental Results}
\subsection{Experimental Setup}
\subsubsection{Downstream Task Datasets}

\paragraph{Wizard of Wikipeida} provides a large-scale dialogue dataset. For each sample, it provides additional knowledge required for each response in the dialogue from Wikipedia. It mainly involves various commonsense knowledge in daily life.

\paragraph{Holl-E} provides a dataset of conversations about specific movie topics. Each conversation in the dataset includes a document that contains information about the corresponding movie, such as the movie plot, movie reviews, and a fact table.

\subsubsection{Large Language Models}
Additionally, we also use some other LLMs in the compared system of our task, including OpenAI's text-davinci-001, text-davinci-002, text-davinci-003, Davinci and gpt-3.5-turbo (ChatGPT).
In our EDIT framework, we choose gpt-3.5-turbo (ChatGPT) as the LLMs used in it.

\subsubsection{Evaluation Metric}
\paragraph{Automatical Evaluation.}
Traditional n-gram metrics such as BLEU (Papineni et al., 2002) and ROUGE (Lin, 2004) are commonly used to measure the similarity between a generated response and a reference or "golden" response.
So we use them to evaluate the generated result of the Question Generation Model.

\paragraph{GPT4 Evaluation.}
The LLM possesses powerful reasoning abilities and an evaluation capacity that aligns with human intentions. As a result, it is extensively employed for downstream task evaluations~\cite{chiang2023vicuna,DBLP:journals/corr/abs-2304-03277,DBLP:journals/corr/abs-2305-11206}. In this work, we utilize GPT4~\cite{DBLP:journals/corr/abs-2303-08774} to assess the generated responses.
To ensure fair evaluation of different models by GPT4, we can input the responses generated by all models into GPT4 and ask it to score each response simultaneously.
The prompt is "Give you a context:\{context\} and some responses: \{Response\_list\}. Please score the response according to whether the answer provide more useful knowledge and give me your score.".

\paragraph{Human Evaluation.}
However, these automatic metrics have limitations in that they only assess the surface-level similarity without considering the semantic coherence, plausibility, and useful information conveyed by the generated response within the given context.
We conducted a human evaluation to assess the generated results of EDIT and other LLMs. 
The human-evaluated Reasonable metric is mainly to evaluate whether the generated response is consistent with the context semantics and to use additional knowledge.

\subsection{Main Results}

On the two downstream tasks of Holl-E and Wizard of Wikipedia, we use the current mainstream LLMs and DEIT to generate responses and evaluate the results.

Table~\ref{tab:Main Result} illustrates the performance of the models on these two tasks. Notably, regarding the Reasonable metric, both the ChatGPT and EDIT(ChatGPT) achieved scores exceeding 70\%, while other models without instruction fine-tuning scored below 50\%. Furthermore, in comparison to ChatGPT, EDIT(ChatGPT) demonstrated remarkable improvements of 21\% and 13.33\% in the two tasks, respectively.
For GPT4 evaluations, both ChatGPT and EDIT(ChatGPT) attained scores surpassing 80\%. On these two downstream tasks, EDIT(ChatGPT) outperformed ChatGPT by 1.0\% and 1.5\%, respectively.
The results of both human evaluation and ChatGPT self-assessment align closely; The output of text-davinci-001 tends to be shorter compared to other models, and this preference for brevity often results in higher scores when evaluated through ChatGPT, although human evaluations may yield varying results.

The results of extensive evaluations, encompassing both ChatGPT Evaluation and Human Evaluation results, conclusively demonstrate that EDIT(ChatGPT) substantially enhanced ChatGPT's performance in dialogue tasks. EDIT(ChatGPT) not only achieves higher scores but also offers information catering to users' intentions, augmenting the overall quality of interactions

\subsection{Ablation Study}

\begin{table}[!htp]
\centering
\scriptsize
\small
\resizebox{0.99\columnwidth}{!}{
    \begin{tabular}{cc}
        \hline
        Models &Wizard of Wikipedia \\
        \hline
        DEIT(ChatGPT) & \textbf{93.50} \\
        -w/o KB's Ans &  91.50\\
        -w/o ChatGPT's Ans & 82.38 \\
        ChatGPT &  72.50 \\
        \hline
         & Holl-E \\
        \hline
        DEIT(ChatGPT) & \textbf{87.62}\\
        -w/o KB's Ans & 86.19\\
        -w/o ChatGPT's Ans & 84.00 \\
        ChatGPT & 74.29 \\
        \hline
    \end{tabular}
}
\caption{The human evaluation result of ablation study for DEIT(ChatGPT).}
\label{tab:Abalation}
\end{table}

To assess the performance of our Question Answering Module of EDIT(ChatGPT) not only achieves higher scores but also offers information catering to users' intentions, we conducted ablation experiments involving Wizard of Wikipedia and Holl-E datasets. Specifically, we deleted the answers of questions provided by ChatGPT (referred to as CG's Ans) and those obtained from a Domain-Specific Knowledge Base(referred to as KB's Ans) on both datasets.

Because, in the ablation experiment, the output results of these models using extra knowledge and the generated responses are relatively rich, it is difficult for ChatGPT to evaluate them well. Therefore we only performed human evaluation.

\begin{figure}[!htp]
\centering
\includegraphics[width=0.99\columnwidth]{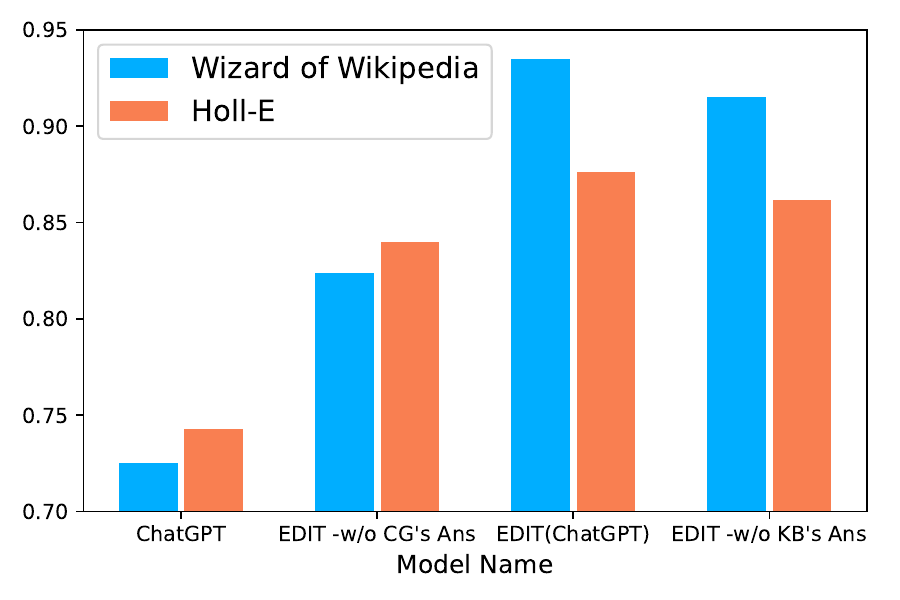}
\caption{The ablation study result of EDIT(ChatGPT).}
\label{figure: Ablation}
\end{figure}

The ablation experiments are shown in Figure~\ref{figure: Ablation} and Table~\ref{tab:Abalation}. On Wizard of Wikipedia dataset and Holl-E dataset, after removing KB's Ans, EDIT(ChatGPT) performance dropped by 2.0\% and 1.43\%; after removing CG's Ans, EDIT(ChatGPT) performance dropped by 11.12\% and 3.62\%.
The experimental results prove that our two Question Answer methods can improve the performance of EDIT(ChatGPT). In particular, adding any type of answer alone can significantly improve the performance of the model, on both datasets, compared with ChatGPT.

\subsection{Question Generation Evaluation}

To enhance LLMs to generate dialogue responses, it is important to identify the users' implicit intentions and use relevant knowledge. In our approach, we employ a Question Generation Model that generates context-related open questions to determine the relevant knowledge. To ensure that the generated questions can indeed enhance the responses and are answerable by LLMs, we conducted an evaluation of these questions.

\subsubsection{Base Models}
For all the Question Generation Model, we use some generative models, such as T5~\cite{2020t5} and BART~\cite{DBLP:journals/corr/abs-1910-13461}, as the base model for fine-tuning, and choose the most suitable model.

\subsubsection{Automatic Metric Results}

\begin{table}[!htp]
\centering
\scriptsize
\small
\resizebox{0.99\columnwidth}{!}{
    \begin{tabular}{cccc}
    \hline
               & \multicolumn{1}{c}{BLEU-1} & \multicolumn{1}{c}{BLEU-2} & \multicolumn{1}{c}{ROUGE-L} \\ \hline
    BART-Base  & 42.24                      & 16.98                      & 13.17                       \\
    BART-Large & 37.08                      & 13.05                      & 11.94                       \\
    T5-Base   & 60.34                      & 35.31                      & 13.44                       \\
    T5-Large   & \textbf{60.72}             & \textbf{35.51}             & \textbf{14.00}              \\ \hline
    \end{tabular}
}
\caption{ The result of the Question Generation Model on Context-Open-Question dataset.
}
\label{tab:Question Model Results}
\end{table}

We assessed the question generation ability of our Question Generation Model of different base models on the test split of Context-Open-Question dataset. Table~\ref{tab:Question Model Results} presents the results, indicating that the Question Generation Model performs well across different base models.

Specifically, the T5 series models exhibit a BLEU-1 score of over 60\%, which is twice as high as the BART series models. Furthermore, the ROUGE-L result of T5-Large surpasses that of BART-Large. Overall, the T5 series demonstrates superior performance in the question generation task compared to other models.

Notably, the performance of our Question Generation Model based on the T5 series remains consistent regardless of the size of the base model. This suggests that even with smaller base models, our Question Generation Model can achieve impressive question generation results while minimizing resource consumption for local deployments.

Taking everything into consideration, we have decided to train the Question Generation Model using T5-Large as the base model.

\begin{table}[!htp]
\centering
\scriptsize
\small
\resizebox{0.9\columnwidth}{!}{
    \begin{tabular}{cc}
    \hline
    Model              & Human Evaluate \\ \hline
    davinci            & 0.20                \\
    text davinci 003 & 0.35               \\
    Question Generation Model     & \textbf{0.94}      \\
    ChatGPT            & 0.52               \\ \hline
    \end{tabular}
}
\caption{The Human Evaluate result of the Quesiont Generation on Context-Open-Question dataset. We report the precision result of Human Evaluation for the generation questions of those four models.
}
\label{tab:Question Model Human Evaluate Result}
\end{table}

\subsubsection{Human Evaluation Results}

In order to evaluate the performance of our Question Model more reasonably, we conduct a human evaluation of the questions generated by Question Generation Model.

Furthermore, we compare our Question Generation Model with other LLMs such as Davinci, Text Davinci 003, and ChatGPT. To do this, we design a specific prompt: "Please generate 5 questions for this context, ensuring that the questions do not involve subjective judgments and focus on well-known objective facts." to use LLMs to generate questions based on context.

Subsequently, we selecte 20 samples and ask each model to generate 100 questions for human evaluation. The results of the evaluation are presented in Table\ref{tab:Question Model Human Evaluate Result}. It is observed that the Question Generation Model achieves an impressive accuracy rate of 94\% for the generated questions, while the other LLMs generally scored lower than 52\%.

These findings serve as strong evidence that our Question Generation Model possesses the ability to design effective questions and assist users in obtaining valuable information from LLMs.

\begin{table*}
\begin{center}
\footnotesize
\resizebox{2.0\columnwidth}{!}{
\begin{tabular}{l}
\hline
\multicolumn{1}{c}{\textbf{Context}}                                                                                                                                                                                                                                                                                                                                                                                                                                                                                                                                                                                                                                                                                                                                     \\
    \begin{tabular}[c]{@{}l@{}}PersonA: I love reading!  It's a means of sharing information and ideas  \\ PersonB: Reading is one of my favorite ways to spend my time. My favorite book series is Harry\\
    \quad \quad \quad \quad Potter by J.K. Rowling. \\ PersonA: Many people love that series! Reading requires continuous practice development and 
    \\ \quad \quad \quad \quad refinement 
    \\ PersonB: So reading can help widen your vocabulary? Are there any other benefits to reading? \end{tabular}                                                                                                                                                                                                                                                                                                                                                                                                                                                                                                                                                                                                                                                                                                                                                             \\ \hline
    \multicolumn{1}{c}{\textbf{Questions Generated by Question Generation Model}}                                                                                                                                                                                                                                                                                                                                                                                                                                                                                                                                                                                                                                                                                                       \\
    \begin{tabular}[c]{@{}l@{}}
    Q1: What are some other ways to spend time besides reading? \\ 
    Q2: How does reading help develop specific skills? \\
    Q3: Are there any specific genres of books that are particularly popular among readers?\\
    Q4: What are some benefits of reading as a means of sharing information and ideas?\\
    Q5: How does reading differ from other forms of entertainment, such as movies or television?\end{tabular}                                                                                                                                                                                                                                                                                                                                                                                                                            \\ \hline
    \multicolumn{1}{c}{\textbf{Responese of ChatGPT}}                                                                                                                                                                                                                                                                                                                                                                                                                                                                                                                                                                                                                                                                                                                        \\
    \begin{tabular}[c]{@{}l@{}}PersonA: Absolutely! Reading not only expands your vocabulary, but it also enhances your co-
    \\mmunication skills. It allows you to explore different writing styles and perspectives, which in\\
    turn improves your own writing abilities. Additionally, reading can boost your empathy and un-
    \\derstanding of others, as you delve into different characters' experiences and emotions. It's 
    \\truly a gateway to knowledge, imagination, and personal growth.\end{tabular}                                                                                                                                                                                                                                                                                                                                              \\ \hline
    \multicolumn{1}{c}{\textbf{Responese of EDIT(ChatGPT)}}                                                                                                                                                                                                                                                                                                                                                                                                                                                                                                                                                                                                                                                                                                                     \\
    \begin{tabular}[c]{@{}l@{}}PersonA: Absolutely! Reading is a wonderful way to expand our knowledge and understanding
    \\ of the world. \sethlcolor{pink}\hl{It allows us to explore different perspectives and gain insights into various subje-}
    \\\sethlcolor{pink}\hl{cts.} Additionally, \sethlcolor{green}\hl{reading enhances critical thinking and analytical skills, as it requires us to ana-}
    \\\sethlcolor{green}\hl{lyze and interpret information.} \sethlcolor{yellow}\hl{It also stimulates our creativity and imagination, allowing us to }
    \\\sethlcolor{yellow}\hl{envision and create vivid worlds and characters in our minds.} Moreover, \sethlcolor{orange}\hl{reading promotes em-}
    \\\sethlcolor{orange}\hl{pathy and understanding by exposing us to different cultures, experiences, and viewpoints.}
    \\Lastly, reading helps improve our communication and language skills, as it exposes us to a wide \\ range of vocabulary and writing styles. Overall, reading is a valuable means of sharing informat-
    \\ion and ideas that offers numerous benefits to individuals.\end{tabular} \\ \hline
    \end{tabular}
}
\end{center}
\caption{The examples from ChatGPT and EDIT(ChatGPT) in Wizard of Wikipedia dataset. The different highlight words represent options related to the benefits of reading.}
\label{tab: Case Study}
\end{table*}

\subsection{Case Study}

We observe that using the EDIT framework can help LLMs respond better.
As shown in Table~\ref{tab: Case Study}, depending on the context, the Question Generation Module of the EDIT can provide questions to cater users' intentions.
Compared with ChatGPT, the responses generated by EDIT(ChatGPT) are not limited to the improvement of writing skills.
Specifically, EDIT(ChatGPT) could provide more options about the benefits of reading, such as 'gain insight', 'enhances critical thinking and analytical skills', 'stimulates creativity and imagination' and 'promote empathy', which may cater to potential users' implicit intentions.
While the response produced by ChatGPT only focuses on communication skills and ignores those potential users' intentions.

\section{Conclusion}
In this study, we present an EDIT framework for the dialogue generation task to address the issue that LLMs may generate plain responses.
Firstly, the Question Generation Module of EDIT generates a variety of questions related to the context of the dialogue as the potential users' implicit intentions. In the Question Answering Module, we interact with LLMs and retrieve in Domain-Specific Knowledge Base to get answers to questions respectively, and we use LLMs to determine the proper answer of question as extra knowledge.
Finally, in Response Generation Module, we use the extra knowledge to enhance the response generated by LLMs.
To evaluate the effectiveness of our approach, we compared the performance of EDIT(ChatGPT) with other LLMs on two downstream dialogue tasks: Holl-E and Wizard of Wikipedia. Our results demonstrate that EDIT achieves SOTA performance on these tasks.

\bibliography{acl2023}
\bibliographystyle{acl2023}

\appendix

\section{Example Appendix}
\label{sec:appendix}

This is a section in the appendix.

\end{document}